\documentclass[10pt,twocolumn,letterpaper]{article}

\usepackage{cvpr}
\usepackage{times}
\usepackage{epsfig}
\usepackage{graphicx}
\usepackage{amsmath}
\usepackage{amssymb}

\usepackage[pagebackref=true,breaklinks=true,letterpaper=true,colorlinks,bookmarks=false]{hyperref}

\cvprfinalcopy

\def\cvprPaperID{63} 

\ifcvprfinal\fi
\begin{document}

\title{Discovering Picturesque Highlights from Egocentric Vacation Videos}

\author{
    Vinay Bettadapura$^{*}$\\
    \texttt{\small vinay@gatech.edu}
  \and
	Daniel Castro$^{*}$\\
    \texttt{\small dcastro9@gatech.edu}
  \and
	Irfan Essa\\
    \texttt{\small irfan@cc.gatech.edu}
  \and
    {\small Georgia Institute of Technology}
  \\
    {\small *These authors contributed equally to this work}
  \\
    \href{http://www.cc.gatech.edu/cpl/projects/egocentrichighlights/}{\small http://www.cc.gatech.edu/cpl/projects/egocentrichighlights/}
}

\maketitle

\begin{abstract}

We present an approach for identifying picturesque highlights from large amounts of egocentric video data. Given a set of egocentric videos captured over the course of a vacation, our method analyzes the videos and looks for images that have good picturesque and artistic properties. We introduce novel techniques to automatically determine aesthetic features such as composition, symmetry and color vibrancy in egocentric videos and rank the video frames based on their photographic qualities to generate highlights. Our approach also uses contextual information such as GPS, when available, to assess the relative importance of each geographic location where the vacation videos were shot. Furthermore, we specifically leverage the properties of egocentric videos to improve our highlight detection. We demonstrate results on a new egocentric vacation dataset which includes 26.5 hours of videos taken over a 14 day vacation that spans many famous tourist destinations and also provide results from a user-study to access our results. 
 
\end{abstract}

\section{Introduction}

Photography is commonplace during vacations. People enjoy capturing the best views at picturesque locations to mark their visit but the act of taking a photograph may sometimes take away from experiencing the moment. With the proliferation of wearable cameras, this paradigm is shifting. A person can now wear an egocentric camera that is continuously recording their experience and enjoy their vacation without having to worry about missing out on capturing the best picturesque scenes at their current location. However, this paradigm results in ``too much data" which is tedious and time-consuming to manually review. There is a clear need for summarization and generation of highlights for egocentric vacation videos.

The new generation of egocentric wearable cameras (i.e. GoPros, Google Glass, etc) are compact, pervasive, and easy to use. These cameras contain additional sensors such as GPS, gyros, accelerometers and magnetometers. Because of this, it is possible to obtain large amounts of long-running egocentric videos with the associated contextual meta-data in real life situations. We seek to extract a series of aesthetic highlights from these egocentric videos in order to provide a brief visual summary of a users' experience.

\begin{figure}
\begin{centering}
\includegraphics[width=1\columnwidth]{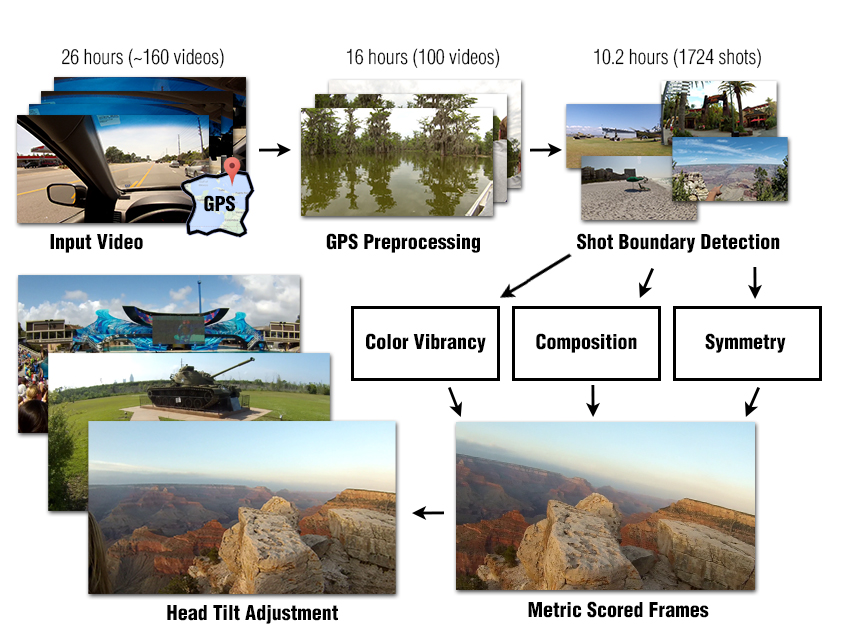}
\par\end{centering}

\caption{\label{fig:block-diagram}Our method generates picturesque summaries and vacation highlights from a large dataset of egocentric vacation videos.}

\vspace{-1.0em}
\end{figure}

Research in the area of egocentric video summarization has mainly focused on life-logging \cite{gemmell2004acm, doherty2008civr} and activities of daily living \cite{fathi2011iccv, pirsiavash2012cvpr, ut2012cvpr}. Egocentric vacation videos are fundamentally different from egocentric daily-living videos. In such unstructured ``in-the-wild" environments, no assumptions can be made about the scene or the objects and activities in the scene. Current state-of-the-art egocentric summarization techniques leverage cues such as people in the scene, position of the hands, objects that are being manipulated and the frequency of object occurrences \cite{fathi2011iccv, ut2012cvpr, fathi2011cvpr, ren2010cvpr, ren2009cvpr, pirsiavash2012cvpr}. These cues that aid summarization in such specific scenarios are not directly applicable to vacation videos where one is roaming around in the world. Popular tourist destinations may be crowded with many unknown people in the environment and contain ``in-the-wild" objects for which building pre-trained object detectors is non-trivial. This, coupled with the wide range of vacation destinations and outdoor and indoor activities, makes joint modeling of activities, actions, and objects an extremely challenging task.

A common theme that exists since the invention of photography is the desire to capture and store picturesque and aesthetically pleasing images and videos. With this observation, we propose to transform the problem of egocentric vacation summarization to a problem of finding the most picturesque scenes within a video volume followed by the generation of summary clips and highlight photo albums. An overview of our system is shown in Figure \ref{fig:block-diagram}. Given a large set of egocentric videos, we show that meta-data such as GPS (when available) can be used in an initial filtering step to remove parts of the videos that are shot at ``unimportant" locations. Inspired by research on exploring high-level semantic photography features in images \cite{luo2008eccv, gooch2001artistic, liu2010cgf,fang2014mm,luo2011iccv,yan2013cvpr}, we develop novel algorithms to analyze the composition, symmetry and color vibrancy within shot boundaries. We also present a technique that leverages egocentric context to extract images with a horizontal horizon by accounting for the head tilt of the user. 

To evaluate our approach, we built a comprehensive dataset that contains 26.5 hours of 1080p HD egocentric video at 30 fps recorded from a head-mounted Contour cam over a 14 day period while driving more than 6,500 kilometers from the east coast to the west coast of the United States. Egocentric videos were captured at geographically diverse tourist locations such as beaches, swamps, canyons, caverns, national parks and at several popular tourist attractions.

\noindent\textbf{Contributions:} This paper makes several contributions aimed at automated summarization of video: (1) We introduce a novel concept of extracting highlight images using photograph quality measures to summarize egocentric vacation videos, which are inherently unstructured. We use a series of methods to find aesthetic pictures, from a large number of video frames, and use location and other meta data to support selection of highlight images. (2) We present a novel approach that accounts for the head tilt of the user and picks the best frame among a set of candidate frames. (3) We present a comprehensive dataset that includes 26.5 hours of video captured over 14 days. (4) We perform a large-scale user-study with 200 evaluators; and (5) We show that our method generalizes to non-egocentric datasets by evaluating on two state-of-the-art photo collections with 500 user-generated and 1000 expert photographs respectively.

\section{Related Work}

We review previous work in video summarization, egocentric analysis and image quality analysis, as these works provide the motivations and foundations for our work.

\noindent\textbf{Video Summarization:} Research in video summarization identifies key frames in video shots using optical flow to summarize a single complex shot \cite{wolf1996icassp}. Other techniques used low level image analysis and parsing to segment and abstract a video source \cite{zhang1997pr} and used a ``well-distributed" hierarchy of key frame sequences for summarization \cite{liu2002eccv}. These methods are aimed at the summarization of specific videos from a stable viewpoint and are not directly applicable to long-term egocentric video.


In recent years, summarization efforts have started focussing on leveraging objects and activities within the scene. Features such as ``informative poses" \cite{caspi2006vc} and ``object of interest", based on labels provided by the user for a small number of frames \cite{liu2010pami}, have helped in activity visualization, video summarization, and generating video synopsis from web-cam videos \cite{pritch2007iccv}.

Other summarization techniques include visualizing short clips in a single image using a schematic storyboard format \cite{goldman2006tog} and visualizing tour videos on a map-based storyboard that allows users to navigate through the video \cite{pongnumkul2008uist}. Non-chronological synopsis has also been explored, where several actions that originally occurred at different times are simultaneously shown together \cite{rav2006cvpr} and all the essential activities of the original video are showcased together \cite{pritch2008pami}. While practical, these methods do not scale to the problem we are adressing of extended videos over days of actvities. 


\noindent\textbf{Egocentric Video Analysis:} Research on egocentric video analysis has mostly focused on activity recognition and activities of daily living. Activities and objects have been thoroughly leveraged to develop egocentric systems that can understand daily-living activities. Activities, actions and objects are jointly modeled and object-hand interactions are assessed \cite{fathi2011iccv, pirsiavash2012cvpr} and people and objects are discovered by developing region cues such as nearness to hands, gaze and frequency of occurrences \cite{ut2012cvpr}. Other approaches include learning object models from egocentric videos of household objects \cite{fathi2011cvpr}, and identifying objects being manipulated by hands \cite{ren2010cvpr, ren2009cvpr}. The use of objects has also been extended to develop a story-driven summarization approach. Sub-events are detected in the video and linked based on the relationships between objects and how objects contribute to the progression of the events \cite{ut2013cvpr}. 

Contrary to these approaches, summarization of egocentric vacation videos simply cannot rely on objects, object-hand interactions, or a fixed category of activities. Vacation videos are vastly different with respect to each other, with no fixed set of activities or objects that can be commonly found across all such videos. Furthermore, in contrast to previous approaches, a vacation summary or highlight must include images and video clips where the hand is not visible and the focus is on the picturesque environment.

Other approaches include detecting and recognizing social interactions using faces and attention \cite{fathi2012cvpr}, activity classification from egocentric and multi-modal data \cite{spriggs2009cvpr}, detecting novelties when a sequence cannot be registered to previously stored sequences captured while doing the same activity \cite{aghazadeh2011cvpr}, discovering egocentric action categories from sports videos for video indexing and retrieval \cite{kitani2011cvpr}, and visualizing summaries as hyperlapse videos \cite{kopf2014tog}.

Another popular area of research and perhaps more relevant is of ``life logging."  Egocentric cameras such as SenseCam \cite{gemmell2004acm} allow a user to capture continuous time series images over long periods of time. Keyframe selection based on image quality metrics such as contrast, sharpness, noise, etc \cite{doherty2008civr} allow for quick summarization in such time-lapse imagery. In our scenario, we have a much larger dataset spanning several days and since we are dealing with vacation videos, we go a step further than image metrics and look at higher level artistic features such as composition, symmetry and color vibrancy.


\noindent\textbf{Image Quality Analysis:} An interesting area of research in image quality analysis is trying to learn and predict how memorable an image is. Approaches include training a predictor on global image features to predict how memorable an image will be \cite{isola2011cvpr} and feature selection to determine attributes that characterize the memorability of an image \cite{isola2011nips}. The aforementioned research shows that images containing faces are the most memorable. However, focusing on faces in egocentric vacation videos causes an unique problem. Since an egocentric camera is always recording, we end up with a huge number of face detections in most of the frames in crowded tourist attractions like Disneyland and Seaworld. To include faces in our vacation summaries, we will have to go beyond face detection and do face recognition and social network analysis on the user to recognize only the faces that the user actually cares about.

The other approach for vacation highlights is to look at the image aesthetics. These include high-level semantic features based on photography techniques \cite{luo2008eccv}, finding good composition for graphics image of a 3D object \cite{gooch2001artistic} and cropping and retargeting based on an evaluation of the composition of the image like the rule-of-thirds, diagonal dominance and visual balance \cite{liu2010cgf}. We took inspiration from such approaches and developed novel algorithms to detect composition, symmetry and color vibrancy for egocentric videos.

\section{Methodology}

Figure \ref{fig:block-diagram} gives an overview of our summarization approach. Let us look at each component in detail.

\subsection{Leveraging GPS Data}

Our pipeline is initiated by leveraging an understanding of the locations the user has traveled throughout their vacation. The GPS data in our dataset is recorded every 0.5 seconds where it is available, for a total of 111,170 points. In order to obtain locations of interest from the data we aggregate the GPS data by assessing the distance of a new point $p_{n}$ relative to the original point $p_{1}$ that the node was created with using the haversine formula which computes the distance between two GPS locations. When the distance is greater than a constant distance $d_{\mathit{max}}$ (defined as 10 km for our dataset) scaled by the speed $s_{p_{n}}$ at which the person was traveling at point $p_{n}$, we create a new node using the new point as the starting location. Lastly, we define a constant $d_{\mathit{min}}$ as the minimum distance that the new GPS point would have to be in order to break off into a new node in order to prevent creating multiple nodes at a single sightseeing location. In summary, a new node is created when $haversine(p_{1}, p_{n}) > s_{p_{n}} * d_{\mathit{max}} + d_{\mathit{min}}$. This formulation aggregates locations in which the user was traveling at a low speed (walking or standing) into one node and those in which the user was traveling at a high speed (driving) into equidistant nodes on the route of travel. The aggregation yields approximately 1,200 GPS nodes in our dataset.

In order to further filter these GPS nodes, we perform a search of businesses / monuments in the vicinity (through the use of Yelp's API) in order to assess the importance of each node using the wisdom of the crowd. The score for each GPS node, $N_{\mathit{score}}$, is given by $N_{\mathit{score}} = \frac{\sum_{l=1}^{L} R_{l} *  r_{l}}{L}$, where $L$ is the number of places returned by the Yelp API in the vicinity of the GPS node $N$, $R_{l}$ is the number of reviews written for each location, and $r_{l}$ is the average rating of each location. This score can then be used as a threshold to disregard nodes with negligible scores and obtain a subset of nodes that represent ``important" points of interest in the dataset.

\subsection{Egocentric Shot Boundary Detection}

Egocentric videos are continuous and pose a challenge in detecting the shot boundaries. In an egocentric video, the scene changes gradually as the person moves around in the environment. We introduce a novel GIST \cite{GIST} based technique that looks at the scene appearance over a window in time. Given $N$ frames $I=<f_{1},f_{2},\ldots,f_{N}>$, each frame $f_{i}$ is assigned an appearance score $\gamma_{i}$ by aggregating the GIST distance scores of all the frames within a window on size $W$ centered at $i$.

\vspace{-1.0em}
\begin{equation}
\gamma_{i}=\frac{\sum_{p=i-\left\lfloor W/2\right\rfloor }^{i+\left\lceil W/2\right\rceil -2}\sum_{q=p+1}^{i+\left\lceil W/2\right\rceil -1}G(f_{p}).G(f_{q})}{[W*(W-1)]/2}
\end{equation}
\vspace{-1.0em}

where $G(f)$ is the normalized GIST descriptor vector for frame $f_{i}$. The score calculation is done over a window to assess the appearances of all the frames with respect to each other within that window. This makes it robust against any outliers within the scene. Since $\gamma_{i}$ is the average of dot-products, its value is between 0 and 1. If consecutive frames belong to the same shot, then their $\gamma$-values will be close to 1. To assign frames to shots, we iterate over $i$ from 1 to $N$ and assign a new shot number to $f_{i}$ whenever $\gamma_{i}$ falls below a threshold $\beta$ (for our experiments, we set $\beta$ = 0.9).

\subsection{Composition}
\label{subsec:composition}

\begin{figure}
\begin{centering}
\includegraphics[width=1\columnwidth]{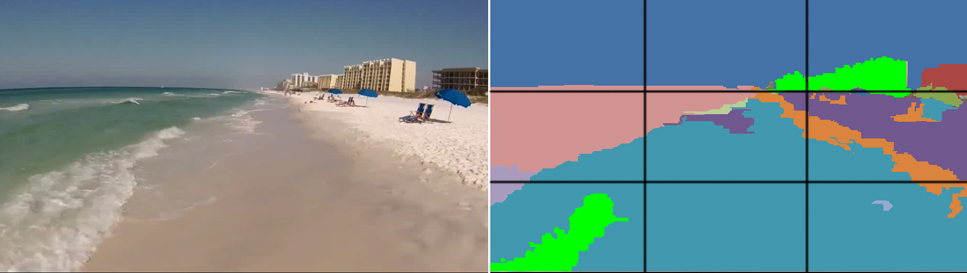}
\par\end{centering}

\caption{\label{fig:composition-segmentation} The left frame shows a highlight detected by our approach. The right frame illustrates the rule-of-thirds grid, overlayed on a visualization of the output of the segmentation algorithm for this particular frame.}

\vspace{-1.0em}
\end{figure}

Composition is one of the characteristics considered when assessing the aesthetics of a photograph \cite{obrador2010role}. Guided by this idea we model composition with a metric that represents the traits of what distinguishes a good composition from a bad composition. The formulation is weighted by a mixture of the average color of specific segments in an image and its distance to an ideal \textbf{rule-of-thirds} composition (see Figure \ref{fig:composition-segmentation}). Our overall results rely on this metric to obtain the highlights of a video clip (see Figure \ref{fig:sample-each} for examples).

\noindent\textbf{Video Segmentation:} The initial step in assessing a video frame is to decompose the frame into cohesive superpixels. In order to obtain these superpixels, we use the public implementation of the hierarchical video segmentation algorithm introduced by Grundmann et. al. \cite{grundmann2010efficient}. We scale the composition score by the number of segments that are produced at a high-level hierarchy (80\% for our dataset) with the intuition that a low number of segments at a high-level hierarchy parameterizes the simplicity of a scene. An added benefit of this parameterization is that a high level of segments can be indicative of errors in the segmentation due to the violation of color constancy which is the underlying assumption of optical flow in the hierarchical segmentation algorithm. This implicitly gets rid of blurry frames. By properly weighting the composition score with the number of segments produced at a higher hierarchy level, we are able to distinguish the visual quality of individual frames in the video.

\noindent\textbf{Weighting Metric:} The overall goal for our composition metric is to obtain a representative score for each frame. First we assess the average color of each segment in the LAB colorspace. We categorize the average color into one of 12 color bins based on their distance, which determines their importance as introduced by Obrador et al. \cite{obrador2010role}. A segment with diverse colors is therefore weighted more heavily than a darker, less vibrant segment. Once we obtain a weight for each segment, we determine the best rule-of-thirds point for the entire frame. This is obtained by computing the score for each of the four points, and simply selecting the maximum. 

\noindent\textbf{Segmentation-Based Composition Metric:} Given $M$ segments for frame $f_{i}$, our metric can be succinctly summarized as the average of the score of each individual segment. The score of each segment is given by the product of its size $s_{j}$ and the weight of its average color $w(c_{j})$, scaled by the distance $d_{j}$ to the rule-of-thirds point that best fits the current frame. So, for frame $f_{i}$, the composition score $S_{\mathit{comp}}^{i}$ is given by:

\vspace{-0.5em}
\begin{equation}
S_{\mathit{comp}}^{i}=\frac{\sum_{j=1}^{M} \frac{s_{j} * w(c_{j})}{d_{j}}}{M}
\end{equation}
\vspace{-1.0em}

\subsection{Symmetry}

Ethologists have shown that preferences to symmetry may appear in response to biological signals, or in situations where there is no obvious signaling context, such as exploratory behavior and human aesthetic response to patterns \cite{enquist1994nature}. Thus, symmetry is the second key factor in our assessment of aesthetics. To detect symmetry in images, we detect local features using SIFT \cite{lowe2004ijcv}, select $k$ descriptors and look for self similarity matches along both the horizontal and vertical axes. When a set of best matching pairs are found, such that the area covered by the matching points is maximized, we declare that a maximal-symmetry has been found in the image. For frame $f_{i}$, the percentage of the frame area that the detected symmetry covers is the symmetry score $S_{\mathit{sym}}^{i}$. 

\subsection{Color Vibrancy}

\begin{figure}
\begin{centering}
\includegraphics[width=0.8\columnwidth]{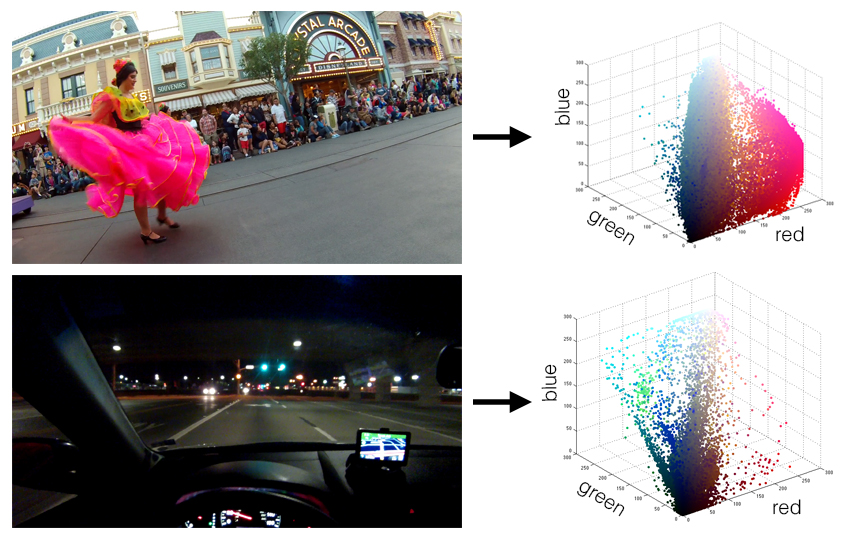}
\par\end{centering}

\caption{\label{fig:color_vibrancy}This visualization demonstrates the difference between a dark frame and a vibrant frame in order to illustrate the importance of vibrancy.}

\vspace{-0.5em}
\end{figure}

\begin{figure}
\begin{centering}
\includegraphics[width=1\columnwidth]{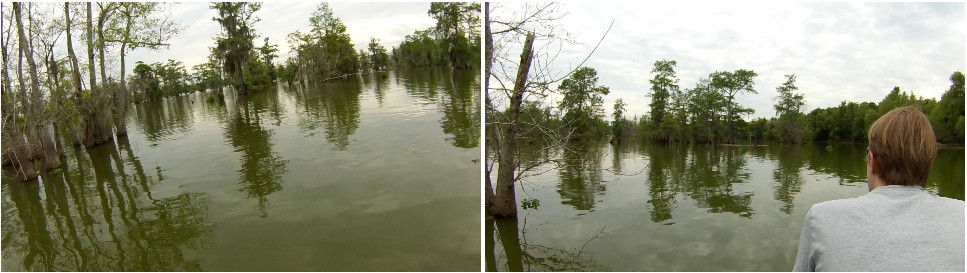}
\par\end{centering}

\caption{\label{fig:head-tilt}Image on left shows a frame with low score on head tilt detection whereas the image on the right has a high score.}

\vspace{-1.0em}
\end{figure}

The vibrancy of a frame is helpful in determining whether or not a given shot is picturesque. We propose a simple metric based on the color weights discussed in Section \ref{subsec:composition} to determine vibrancy. This metric is obtained by quantizing the colors of a single frame into twelve discrete bins and scaling them based on the average distance from the center of the bin. This distance represents the density of the color space for each bin which is best appreciated by the visualization in Figure \ref{fig:color_vibrancy}. The vibrancy score for frame $f_{i}$ is given by:

\vspace{-1.5em}
\begin{equation}
S_{\mathit{vib}}^{i}=\sum_{j=1}^{n_{b}} \frac{w(c_{j}) * b_{\mathit{size}}}{b_{\mathit{dist}}}
\end{equation}
\vspace{-1.0em}

where $n_{b}$ is the number of color bins (12 in our case), $w(c_{j})$ is the color weight, $b_{\mathit{size}}$ is the bin size (number of pixels in the bin) and $b_{\mathit{dist}}$ is the average distance of all the pixels to the actual bin color.

\subsection{Accounting For Head Tilt}

Traditional approaches on detecting aesthetics and photographic quality in images take standard photographs as input. However, when dealing with egocentric video, we also have to account for the fact that there is a lot of head motion involved. Even if we get high scores on composition, symmetry, and vibrancy, there is still a possibility that the head was tilted when that frame was captured. This diminishes the aesthetic appeal of the image.

While the problem of horizon detection has been studied in the context of determining vanishing points, determining image orientations and even using sensor data on phones and wearable devices \cite{wang2012ubicomp}, it still remains a challenging problem. However, in the context of egocentric videos, we approach this by looking at a time window around the frame being considered. The key insight is that while a person may tilt and move his head at any given point in time, the head remains straight \textit{on average}. With this, we propose a novel and simple solution to detect head tilt in egocentric videos. We look at a window of size $W$ around the frame $f_{i}$ and average all the frames in that window. If $f_{i}$ is similar to average frame, then the head tilt is deemed to be minimal. For comparing $f_{i}$ to the average image, we use the SSIM metric \cite{hore2010image} as the score $S_{\mathit{head}}^{i}$ for frame $f_{i}$. Figure \ref{fig:head-tilt} shows two sample frames with low and high scores.

\subsection{Scoring and Ranking}

We proposed four different metrics (composition, symmetry, vibrancy, head tilt) for assessing aesthetic qualities in egocentric videos. Composition and symmetry are the foundation of our pipeline, and vibrancy and head tilt are metrics for fine-tuning our result for a picturesque output. The final score for frame $f_{i}$ is given by:

\vspace{-0.5em}
\begin{equation}
S_{\mathit{final}}^{i}=S_{\mathit{vib}}^{i}*(\lambda_{1}*S_{\mathit{comp}}^{i}+\lambda_{2}*S_{\mathit{sym}}^{i})
\end{equation}
\vspace{-1.0em}

Our scoring algorithm assesses all of the frames based on a vibrancy weighted sum of composition and symmetry (empirically determined as ideal: $\lambda_{1} = 0.8$, $\lambda_{2} = 0.2$). This enables us to obtain the best shots for a particular video. Once we have obtained $S_{\mathit{final}}^{i}$, we look within its shot boundary to find the best $S_{\mathit{head}}^{i}$ that depicts a well composed frame.

\section{Egocentric Vacation Dataset}

\begin{figure}
\begin{centering}
\includegraphics[width=1\columnwidth]{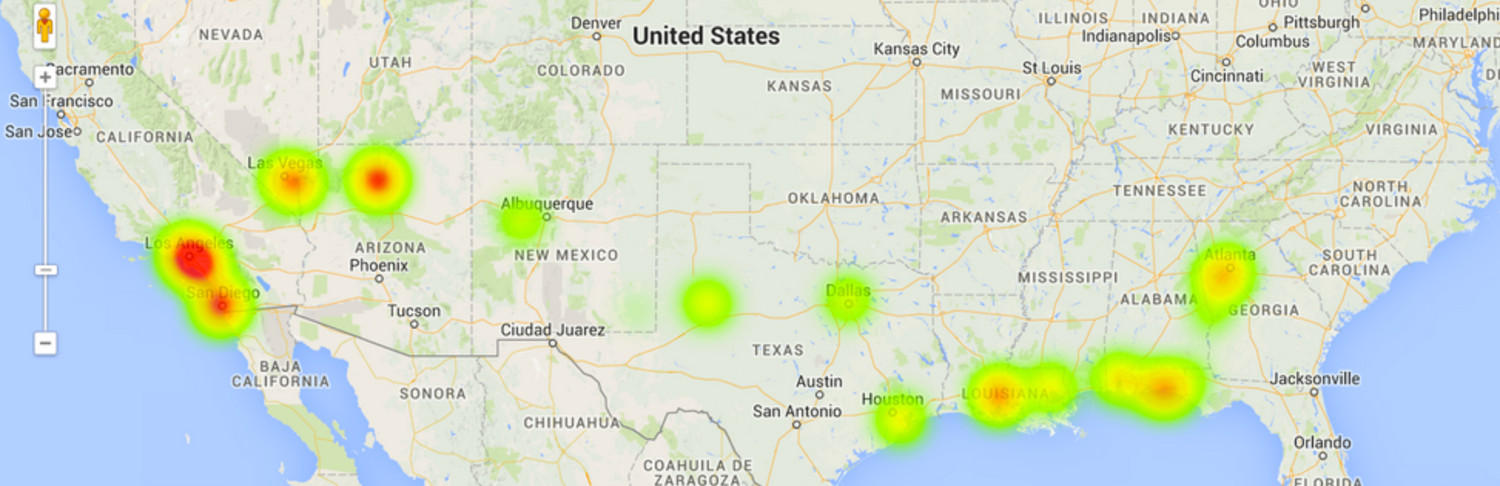}
\par\end{centering}

\caption{\label{fig:dataset-map}A heatmap showing the egocentric data collected while driving from the east coast to the west coast of the United States over a period of 14 days. Hotter regions on the map indicate the availability of larger amounts of video data.}

\vspace{-0.5em}
\end{figure}

To build a comprehensive dataset for our evaluation, we drove from the east coast to the west coast of the United States over a 14 day period with a head-mounted Contour cam and collected egocentric vacation videos along with contextual meta-data such as the GPS, speed and elevation. Figure \ref{fig:dataset-map} shows a heatmap of the locations where data was captured. Hotter regions indicate availability of more data.

\begin{figure}
\begin{centering}
\includegraphics[width=1\columnwidth]{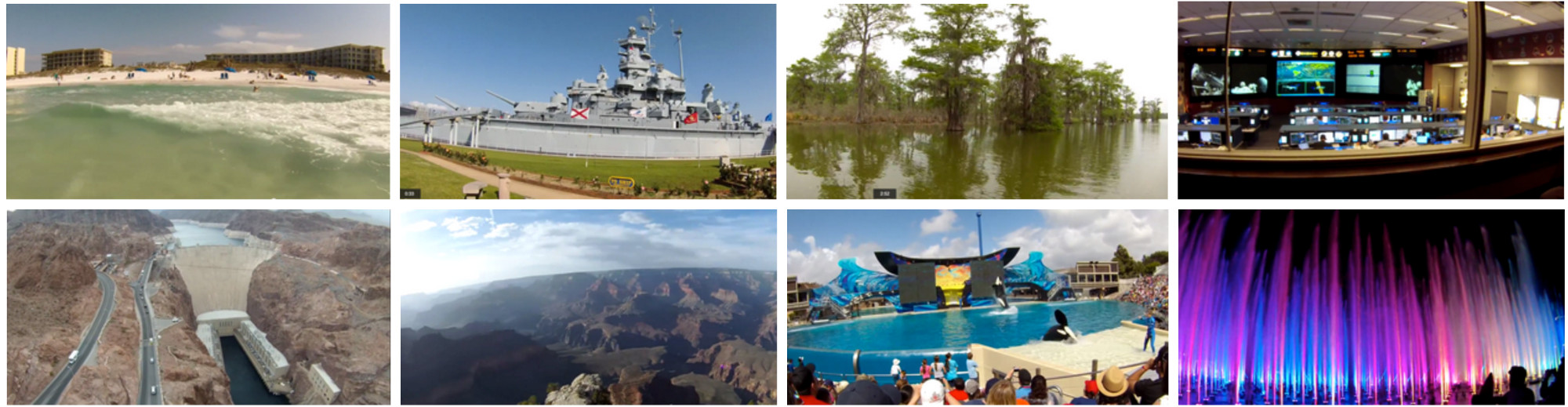}
\par\end{centering}

\caption{\label{fig:dataset-images}Sample frames showing the diversity of our egocentric vacation dataset. The dataset includes over 26.5 hours of HD egocentric video at 30 fps.}
\vspace{-0.5em}
\end{figure}

The dataset has over 26.5 hours of 1080p HD egocentric video (over 2.8 million frames) at 30 fps. Egocentric videos were captured at geographically diverse locations such as beaches, swamps, canyons, national parks and popular tourist locations such as the NASA Space Center, Grand Canyon, Hoover Dam, Seaworld, Disneyland, and Universal Studios. Figure \ref{fig:dataset-images} shows a few sample frames from the dataset. To the best of our knowledge, this is the most comprehensive egocentric dataset that includes both HD videos at a wide range of locations along with a rich source of contextual meta-data.

\begin{figure*}
\begin{centering}
\includegraphics[width=1\textwidth]{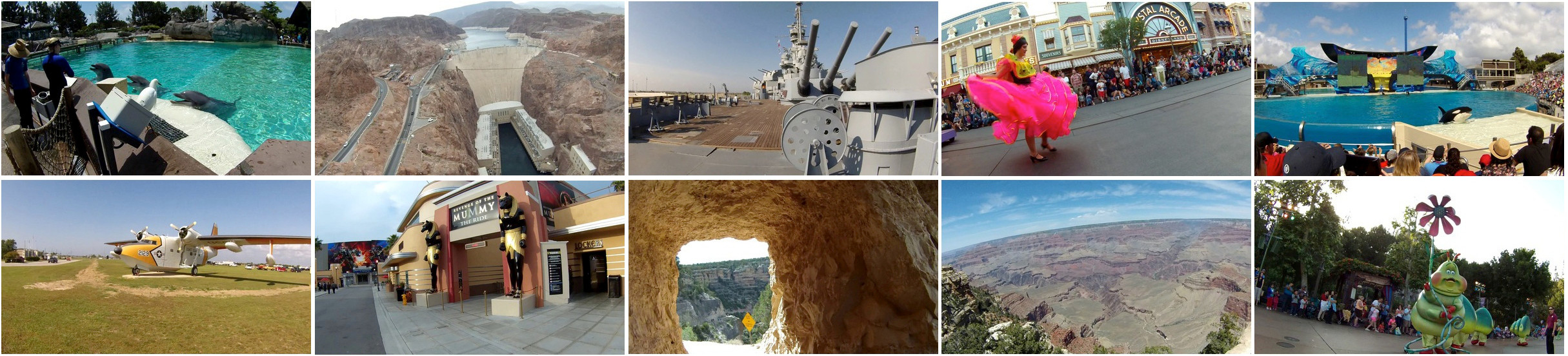}
\par\end{centering}

\caption{\label{fig:sample-all}10 sample frames that were ranked high in the final output. These are the types of vacation highlights that our system outputs.}
\vspace{-1.0em}
\end{figure*}

\section{Evaluation}

\begin{figure}
\begin{centering}
\includegraphics[width=1\columnwidth]{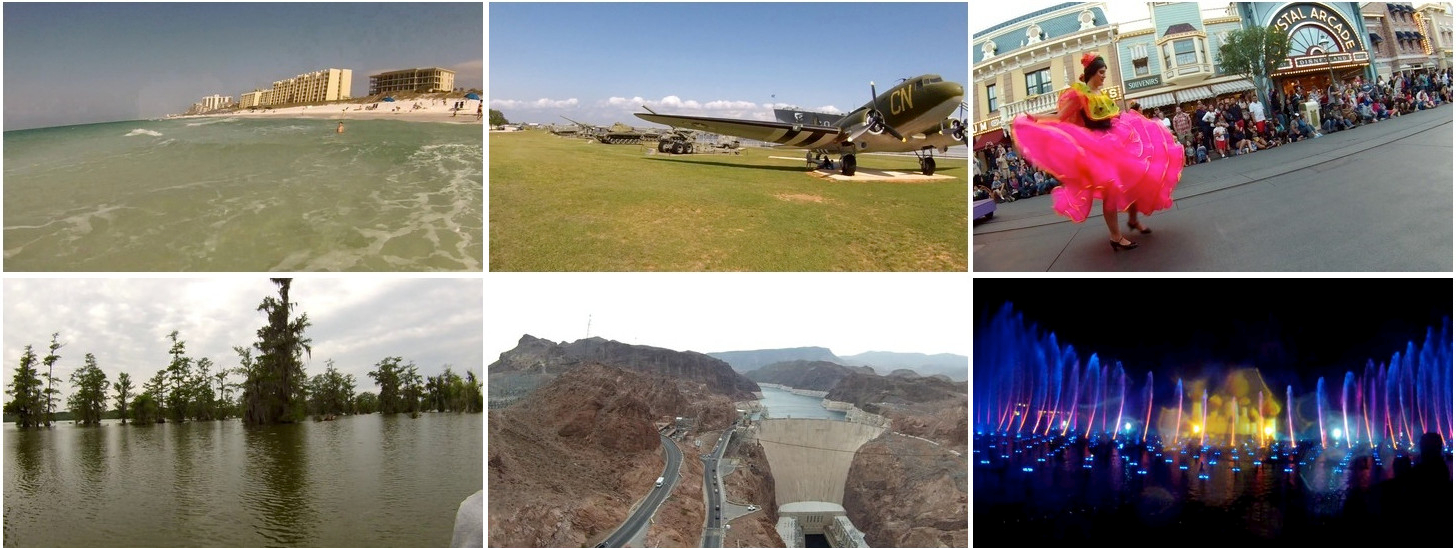}
\par\end{centering}

\caption{\label{fig:sample-each}Top row shows 3 samples frames that were ranked high in composition alone and the bottom row shows 3 sample frames that were ranked high in symmetry alone.}
\vspace{-1.0em}
\end{figure}

We performed tests on the individual components of our pipeline in order to assess the output of each individual metric. Figure \ref{fig:sample-each} shows three sample images that received high scores in composition alone and three sample images that received high scores in symmetry alone (both computed independent of other metrics). Based on this evaluation, which gave us an insight into the importance of combining frame composition and symmetry, we set $\lambda_1=0.8$ and $\lambda_2=0.2$. Figure \ref{fig:sample-all} depicts 10 sample images that were highly ranked in the final output album of 100 frames. In order to evaluate our results, which are inherently subjective, we conduct A/B testing on two baselines with a notable set of subjects on Amazon Mechanical Turk. 

\subsection{Study 1 - Geographically Uniform Baseline}

\begin{figure}
\begin{centering}
\includegraphics[width=0.75\columnwidth]{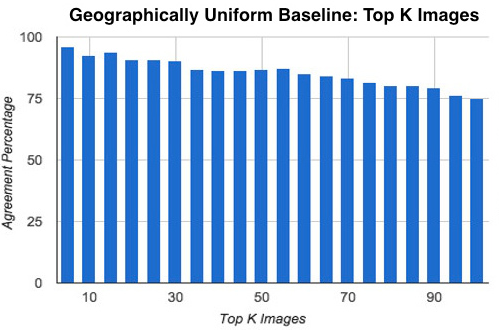}
\par\end{centering}

\caption{\label{fig:results_baseline_1}This figure demonstrates the agreement percentage for the top k images of our pipeline. For instance, for the top 50\% images, we have an agreement percentage of 86.67\%. This represents the number of users in our study that believed that our images were more picturesque than the baseline.}

\vspace{-1.0em}
\end{figure}

Our first user study consists of 100 images divided over 10 Human Intelligence Tasks (HIT) for 200 users (10 image pairs per HIT). To get good quality, we required participants to have an approval rating of 95\% and a minimum of 1000 approved HITs. The HITs took an average time of 1 minute and 6 seconds to complete and the workers were all rewarded \$0.06 per HIT. Due to the subjective nature of the assessment, we opted to approve and pay all of our workers within the hour.

\noindent\textbf{Baseline:} For this baseline we select $x$ images that are equally distributed across the GPS data of the entire dataset. This was performed by uniformly sampling the GPS data and selecting the corresponding video for that point. After selecting the appropriate video we select the closest frame in time to the GPS data point. We were motivated to explore this baseline due to the nature of the dataset (data was collected from the East to the West coast of the United States). The main benefit of this baseline is that it properly represents the locations throughout the dataset and is not biased by the varying distribution of videos that can be seen in the heatmaps in Figure \ref{fig:dataset-map}.

\noindent\textbf{Experiment Setup:} The experiment had a very straightforward setup. The title of the HIT informed the user of their task, ``Compare two images, click on the best one.". The user was presented with 10 pairs of images for each task. Above each pair of images, the user was presented with detailed instructions, ``Of these two (2) images, click which one you think is better to include in a vacation album.".  The left / right images and the order of the image pairs were randomized for every individual HIT in order to remove bias. Upon completion the user was able to submit the HIT and perform the next set of 10 image comparisons. Every image the user saw within a single HIT and the user study was unique and therefore not repeated across HITs. The image pair was always the same, so users were consistently comparing the same pair (albeit with random left / right placement). Turkers were incredibly pleased with the experiment and we received extensive positive feedback on the HITs.

\noindent\textbf{Results:}  Figure \ref{fig:results_baseline_1} demonstrates the agreement percentage of the user study from the top five images to the top 100, with a step size of 5. For our top 50 photo album, we obtain an agreement percentage from 200 turkers of 86.67\%. However, for the top 5-30 photos, we obtain an agreement of greater than 90\%. We do note the inverse correlation between album size and agreement which is due to the increasing prevalence of frames taken from inside the vehicle while driving and the general subjectiveness of vacation album assessment.

\subsection{Study 2 - Chronologically Uniform Baseline }

\begin{figure}
\begin{centering}
\includegraphics[width=0.75\columnwidth]{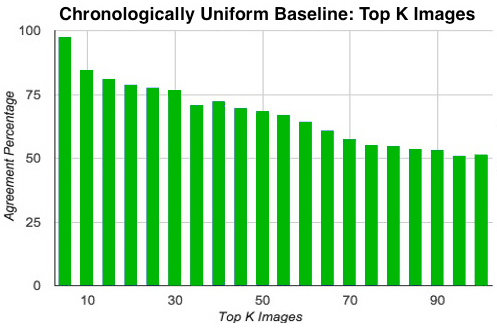}
\par\end{centering}

\caption{\label{fig:results_baseline_2}This figure demonstrates the average agreement percentage among 50 master turkers for our top k frames. For instance, for our top 50 frames, we obtain an agreement percentage of 68.68\%.}

\vspace{-0.5em}
\end{figure}

Our second user study consists of 100 images divided over 10 HITs (10 per HIT) for 50 Master users (Turkers with demonstrated accuracy). These HITs took an average time of 57 seconds to complete and the workers were all rewarded \$0.10 per HIT. 

\begin{figure}
\begin{centering}
\includegraphics[width=1\columnwidth]{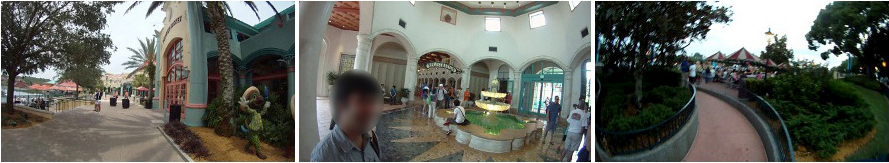}
\par\end{centering}

\caption{\label{fig:comparison}Three sample highlights from the Egocentric Social Interaction dataset \cite{fathi2012cvpr}}

\vspace{-1.0em}
\end{figure}

\noindent\textbf{Baseline:} In this user study we developed a more challenging baseline in which we do not assume an advantage by using of GPS data. Our pipeline and the chronological uniform baseline are both given clips after the GPS data has parsed out the ``unimportant" locations. The baseline uniformly samples in time across the entire subset of videos and selects those frames for comparison. We do note that the distribution of data is heavily weighted on important regions of the dataset where a lot of data was collected, which adds to the bias of location interest and the challenging nature of this baseline.

\noindent\textbf{Experimental Setup:} The protocol for the chronologically uniform baseline was identical. Due to the difficult baseline, we increase the overall requirements for Mechanical Turk workers and allowed only ``Masters" to work on our HITs. We decreased our sample size to 50 Masters due to the difficulty of obtaining turkers with Masters certification. The title and instructions from the previous user study were kept identical along with the randomization of the two images within a pair, and the 10 pairs within a HIT.

\noindent\textbf{Results:} For the top 50 images, we obtain an agreement percentage of 68.67\% (See Figure \ref{fig:results_baseline_2}). We once again note the high level of agreement for the top 5 images, 97.7\% agree the images belong in a vacation photo album. These results reinforce our pipeline as a viable approach to determining quality frames from a massive dataset of video. We also note the decrease in accuracy beyond 50 images, in which the agreement percentage between turkers reaches 51.42\% for all the top 100 images. We believe this is due to the difficulty of the baseline, and the hard constraint on the number of quality frames in interesting locations that are properly aligned and unoccluded.

\begin{figure}
\begin{centering}
\includegraphics[width=0.95\columnwidth]{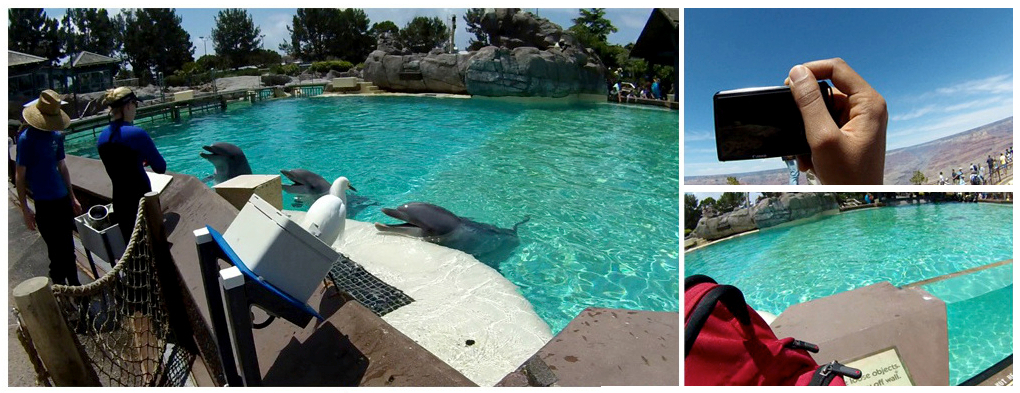}
\par\end{centering}

\caption{\label{fig:assessment}Left: 95\% agreement between turkers that they would include this picture in their vacation album. Top Right: 62\% agreement. Bottom Right: 8\% agreement.}

\vspace{-1.5em}
\end{figure}

\subsection{Assessing Turker Agreement}

In Figure \ref{fig:assessment}, we can see three output images that had varying levels of agreement percentages between turkers. The left image with 95\% agreement between Turkers is a true-positive, which is a good representation of a vacation image. The top-right and bottom-right images are two sample false positives that were deemed to be highlights by our system. These received 62\% and 8\% agreement respectively. We observe false positives when the users' hand breaches the rule of thirds' region (like the top-right image), thereby firing erroneous high scores in composition. Also, random bright colored objects (like the red bag in front of the greenish-blue water in the bottom-right image) resulted in high scores on color vibrancy.

\subsection{Generalization on Other Datasets}

\noindent\textbf{Egocentric Approaches:} Comparing our approach to other egocentric approaches is challenging due to the applicability of other approaches to our dataset. State-of-the-art techniques on egocentric videos such as \cite{ut2012cvpr,ut2013cvpr} focus on activities of daily living and rely on detecting commonly occurring objects, while approaches such as \cite{fathi2011iccv,fathi2011cvpr} rely on detecting hands and their relative position to the objects within the scene. In contrast, we have in-the-wild vacation videos without any predefined or commonly occurring object classes. Other approaches, such as \cite{gygli2014eccv}, perform superframe segmentation on the entire video corpus which does not scale to 26.5 hours of egocentric videos. Further, \cite{fathi2012cvpr} uses 8 egocentric video feeds to understand social interactions which is distinct from our dataset and research goal. However, we are keen to note that the Social Interactions dataset collected at Disneyland by \cite{fathi2012cvpr} was the closest dataset we could find to resemble a vacation dataset due to its location. We ran our pipeline on this dataset, and our results can be seen in Figure \ref{fig:comparison}. The results are representative of vibrant, well-composed, symmetric shots which reinforce the robustness of our pipeline. We do note that these results are obtained without GPS preprocessing which was not available / applicable to that dataset.

\begin{figure}
\begin{centering}
\includegraphics[width=1\columnwidth]{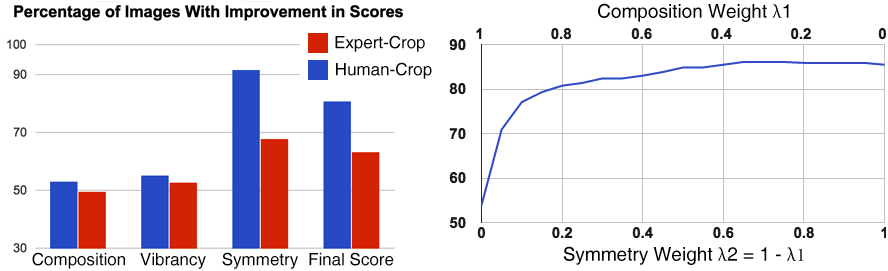}
\par\end{centering}

\caption{\label{fig:improvement}Left: Percentage of images with an increase in the final score for both the Human-Crop dataset \cite{fang2014mm} and Expert-Crop dataset \cite{luo2011iccv,yan2013cvpr}. Right: Percentage of images in the Human-Crop dataset with an increase in the final score as a function of the composition and symmetry weights.}

\vspace{-0.5em}
\end{figure}

\begin{figure}
\begin{centering}
\includegraphics[width=1\columnwidth]{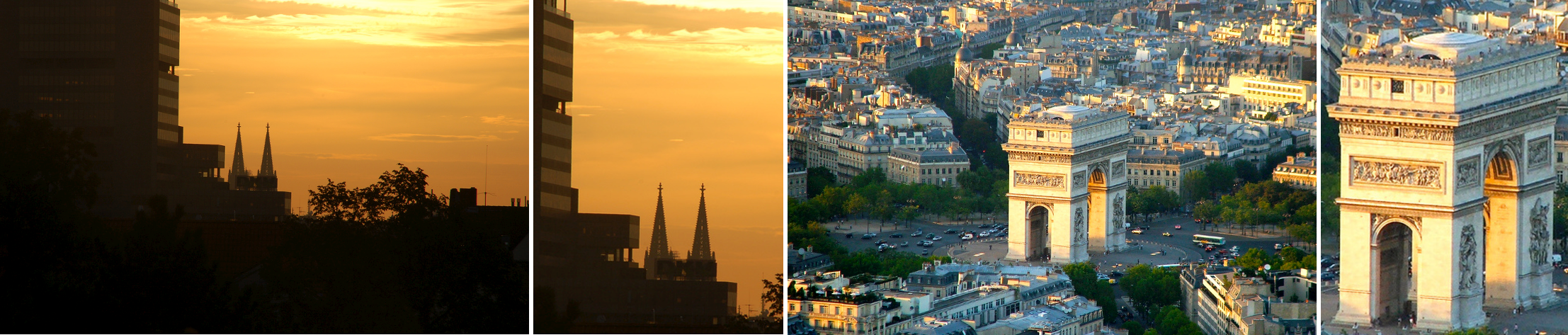}
\par\end{centering}

\caption{\label{fig:improved_symetry}Two examples of the original images and the images cropped by expert photographers. Note the improvement in the overall symmetry of the image.}

\vspace{-1.5em}
\end{figure}

\noindent\textbf{Photo Collections:} In order to analyze the external validity of our approach on non-egocentric datasets, we tested our methodology on two state-of-the-art photo collection datasets. The first dataset \cite{fang2014mm} consists of 500 user-generated photographs. Each image was manually cropped by 10 Master users on Amazon Mechanical Turk. We label this dataset the ``Human-Crop dataset". The second dataset \cite{luo2011iccv,yan2013cvpr} consists of 1000 photographs taken by amateur photographers. In this case, each image was manually cropped by three expert photographers (graduate students in art whose primary medium is photography). We label this dataset the ``Expert-Crop dataset". Both datasets have aesthetically pleasing photographs spanning a variety of image categories, including architecture, landscapes, animals, humans, plants and man-made objects.

To assess our metrics effectiveness we ran our pipeline (with $\lambda_1=0.8$ and $\lambda_2=0.2$) on both the original uncropped images and the cropped images provided by the human labelers. Since the cropped images are supposed to represent an aesthetic improvement, our hypothesis was that we should see an increase in our scoring metrics for the cropped images relative to the original shot. For each image in the dataset, we compare the scores of each of the cropped variants (where the crops are provided by the labelers) to the scores of the original image. The scores for that image are considered an improvement only if we see an increase in a majority of its cropped variants. Figure \ref{fig:improvement} (left) shows the percentage of images that saw an improvement in each of the four scores: composition, vibrancy, symmetry and the overall final score. We can see that the final score was improved for 80.74\% of the images in the Human-Crop dataset and for 63.28\% of the images in the Expert-Crop dataset.

We are keen to highlight that the traditional photography pipeline begins with the preparation and composition of the shot in appropriate lighting and finishes with post-processing the captured light using state-of-the-art software. Hence, the cropping of the photograph is a sliver of the many tasks undertaken by a photographer. This is directly reflected in the fact that we do not see a large increase in the composition and vibrancy scores for the images as those metrics are somewhat irrespective of applying a crop window within a shot that has already been taken. The task of cropping the photographs has its most direct effect in making the images more symmetrical. This is reflected in the large increase in our symmetry scores. Two examples of this can be seen in Figure \ref{fig:improved_symetry}. To test this hypothesis further, we ran an experiment on the Human-Crop dataset where we varied the composition weight $\lambda_1$ between 0 and 1 and set the symmetry score $\lambda_2=1-\lambda_1$. From Figure \ref{fig:improvement} (right), we can see that the percentage of images that saw an increase in the final score increases as $\lambda_1$ (the composition weight) decreases and $\lambda_2$ (the symmetry weight) increases. Also note that we see a larger improvement in our scores for the Human-Crop dataset when compared to the Expert-Crop dataset. This behavior is representative of the fact that the Expert-Crop dataset has professional photographs that are already very well-composed (and cropping provides only minor improvements) when compared to the Human-Crop dataset that has user-generated photographs where there is more scope for improvement with the use of a simple crop.

\section{Conclusion}

In this paper we presented an approach that identifies picturesque highlights from egocentric vacation videos. We introduce a novel pipeline that considers composition, symmetry and color vibrancy as scoring metrics for determining what is picturesque. We reinforce these metrics by accounting for head tilt using a novel technique to bypass the difficulties of horizon detection. We further demonstrate the benefits of meta-data in our pipeline by utilizing GPS data to minimize computation and better understand the places of travel in the vacation dataset. We exhibit promising results from two user studies and the generalizability of our pipeline by running experiments on two other state-of-the-art photo collection datasets.

\newpage

{
\small\bibliographystyle{ieee}
\bibliography{egocentric}
}

\end{document}